\documentclass[journal]{article}
\usepackage{graphicx} 
\usepackage{amsmath}
\usepackage{booktabs} 
\title{Attractor FCM}
\author{Alexis Kafantaris}
\date{April 2026}

\begin{document}

\maketitle
\section*{Abstract}
In this paper an attractor FCM is created, tested, and analyzed. This FCM is neither a hebbian based nor agentic, nor a hybrid; it rather is a gradient descent based, physics constrained, Jacobian version of an FCM. Moreover, this model has several quirks; it uses residual memory, back propagation through time, and a fixed point anchor that is recursively implemented to update its weights. The residuals update the recursive part without losing the system memory. The model's anchor enables it to converge in a fixed point for which back propagation through time unrolls it and ensures that the error minimization is for an accurate gradient. Furthermore, a new learning algorithm is utilized. The Newton's method finds the system's fixed point attractor and then gradient descend is adaptively changing the landscape; an adaptive term is used to directly manipulate the weights through the attractor dynamics. As the adaptive term changes, the descent through the landscape is constantly adjusting according to sigmoid saturation, and that prevents premature convergence to a local minimum. Lastly, the updates are filtered by causal mask that informs the network about the physics, respecting the initial expert based opinions, for which model reduces the error to the target in an efficient way.
\pagebreak
\section{Introduction}
Fuzzy cognitive map learning methods is an interesting topic that has been somewhat stale \cite{stach2010}. Although there are many programs and learning methods, a better approach is still lacking \cite{tsadiras2008}. Some methods have convergence advantages, while others have interpretability. For example, Hebbian \cite{hebb1949} based methods converge faster but are not interpretable. On the other hand, an agentic setup might lack in convergence but it is interpretable \cite{eiben2015}. Ideally, each setup is meant to be used under specific conditions, but one does not want to make that choice; or rather one prefers a fuzzy cognitive map that is holistically better.

Moreover, fuzzy cognitive maps are used for classification, prediction, and simulation \cite{kosko1997}. These are already key distinctions which more or less shape the architecture. Again, the different versions of fuzzy cognitive maps as well as their learning methods are shaped to address the key challenges \cite{papageorgiou2011}. Simulation, for example, accepts a target and then feeds it into the fuzzy cognitive map network. The network, attempting to minimize the error, adjusts the weights which represent concept correlation \cite{kosko1986}; then by assessing these concepts one can determine the simulation outcome as the mathematical equilibrium that the system settles into, assuming that the system does not fall into chaos.

Furthermore, for prediction, previous inputs are used and given these, new targets are evaluated \cite{elman1990}. The important idea is that the fuzzy cognitive map will be adjusted to minimize the error, so in turn it will explain the way that it made its deductions. Lastly, for classification the idea is the same; however, all these different applications require a different setup \cite{tsadiras2008}. Nonetheless, the tradeoffs are simple and make intuitive sense; assuming one wants a simulation, he only cares about the output targets and so wants fast convergence. Therefore a Hebbian FCM is a good choice in a computational sense \cite{hebb1949}; however, this lacks in interpretability, and so trying to read the map will make no sense.

Last but not least there are other networks as well from which one can draw inspiration. By assesing deep equilibrium models and hopefield networks one finds that there are other methods like the back propagation through time which are used. In turn, these methods could be implemented in a fuzzy cognitive map leveraging the math and creating hybrids, or more pure forms of the fuzzy cognitive maps. Hence, achieving better convergence and other interesting properties which are examined in the attractor FCM.
\section{Related Works}

A fuzzy cognitive map is a signed digraph; a symbolic network with concepts as nodes and fuzzy relationships as edges. By using fuzzy relations as edges it becomes interpretable, e.g. concept C1 affects to concept C2 by 0.2, edge C1C2 is 0.2, or concept C3 affects concept C2 by negative 0.5, edge C3C2 is 0.5. The FCM is a shallow recurrent neural network that instead of calculating random weights, it calculates weights as system dynamics \cite{kosko1986}. This way, one can simulate the system, and through simulation determine the mathematical equilibrium \cite{kosko1997}. 
The FCM Equations are as follows:

\[
H_{t+1} = \tanh(H_t W), \quad H_0 = H_{\text{initial}}
\]

$$\;H_T = \Phi \circ \Phi \circ \cdots \circ \Phi (H_0)\;$$

\[
E = H_{\text{target}} - H_T
\]

\[
W  =   W + \eta \, H_T \otimes E
\]

According to the FCM literature the FCM learning methods that are suggested are hebbian based \cite{hebb1949}, population based \cite{dorigo2004,eiben2015} and hybrids \cite{papageorgiou2011}. For all of these there are advantages regarding explainability, convergence, and other properties \cite{stach2010}. Moreover, due to the scarcity of algorithms in that domain there is a gap. The existing algorithms are somewhat limited in strengths as well and depending on the algorithm there are several advantages and disadvantages \cite{tsadiras2008}. Generally, there are other similar ideas to an FCM like the DEQ networks \cite{bai2019deq}, which can be viewed as optimization frameworks of a fixed point; and there are Hopfield networks \cite{hopfield1982,hopfield1984}, that use energy minimization to make inferences, and may implement interesting dynamics. These networks all use ideas that should apply and improve an FCM, however there is nothing formal established yet.

The gradient descent algorithm is a mathematical tool for finding the solution. Using the derivative of the feed forward equations the gradient descent can find the best direction for back propagating the weights. In other words, by moving towards the direction of the gradient time the program reduces the error of the target by a step epsilon; as the error decreases the model is closer to the target, which makes the gradient descent a very efficient program. However, having to rely on a step each time for that process means that the algorithm can get stuck at a local minima that is deeper than a step. On the other hand, too large of a step and the algorithm gets unstable.

This algorithm is inspired by equilibrium propagation networks \cite{scellier2017}, and from the gradient descent component, by improving on a gradient descent based method the gap is addressed; using lyapunov stability \cite{lyapunov1992}, the energy is minimized and then the inputs are denoised under a condition of contraction \cite{bengio2013}. In addition, a structural mask is applied on the network, so this algorithm retains a hundred percent of its initial structure. The structure represents the physics of a system, the experts opinion is reflected through these interconnections; respecting the physics and minimizing the targets using Jacobian gradient descent is an excellent approach. Additionally, the system also has a denoising properties. Hence, the system for a given input the system acts like a catalyst, preserving the graph topology while also optimizing for the specific targets.

\section{Methodology}

Here, the attractor FCM model is created; the key components are established and a new learning method is formalized. Then, convergence and denoisification are proved using Lipschitz contraction and Banach fixed point theorem (Contraction mapping) 
\subsection*{Attractor FCM: System Definitions}
\[
M = (W_{\text{initial}} \neq 0)
\]
\[
S = \operatorname{sign}(W_{\text{initial}})
\]
\[
H_{T_0} = H_{\text{curr}}
\]
\[
\tilde{H}_{t+1} = \sigma\left(H_t W + H_t\right)
\]
\[
H_{t+1} = (1 - \alpha) H_t + \alpha \tilde{H}_{t+1}, \quad \alpha \in (0,1]
\]
At this point, residual recursion memory is used to smoothen the transitions \cite{elman1990}. The attractor state $H_T$ is reached after $T$ iterations from the initial state:
\[
\Phi(H_{t}) = (1 - \alpha)H_{t}+ \alpha \, \sigma\left(H_{t}W + H_{t}\right)
\]
\[
H_{T} =
\underbrace{\Phi \circ \Phi \circ \cdots \circ \Phi}_{T \text{ times}}
(H_{0})
\]

\vspace{0.3cm}
To preserve the causal and structural integrity of the system, the weight matrix is constrained such that:
\[
W_{ij} =
\begin{cases}
W_{ij} & \text{if } M_{ij} = 1 \\[6pt]
0 & \text{if } M_{ij} = 0
\end{cases}
\]

\vspace{0.5cm}
\subsection*{J-GD: Jacobian Gradient Descent}

Rather than propagating gradients through $T$ unrolled steps, J-GD finds the
true fixed point $H^*$ of the FCM via Newton's method from the attractor,
in other words it updates $W$ directly from the fixed-point geometry \cite{bai2019deq,scellier2017}.

\vspace{0.3cm}
\textbf{Step 1 — Newton Fixed Point.}
The fixed point $H^*$ satisfies the self-consistency equation \cite{hopfield1982,ramsauer2021hopfield}:
\[
H^* = \sigma\!\left(H^*(W + I)\right)
\]
Define the residual $F(H) = \sigma(H(W+I)) - H$. Newton's method iterates:
\[
H^{(k+1)} = H^{(k)} - J_F^{-1}\, F\!\left(H^{(k)}\right)
\]
where the Jacobian is:
\[
J_F = \operatorname{diag}\!\left(\sigma'\!\left(H^{(k)}(W+I)\right)\right)(W + I) - I
\]
and $\sigma'(z) = \sigma(z)(1 - \sigma(z))$ is the sigmoid derivative.
Iteration proceeds until $\|F(H^{(k)})\| < \varepsilon$.

\vspace{0.3cm}
\textbf{Step 2 — Adaptive Scale $\lambda$.}
The scale factor $\lambda$ is derived from the ratio of the current BPTT
gradient norm to the reference norm at initialisation:
\[
\lambda = \operatorname{clip}\!\left(\frac{1}{\|\nabla_W\| / \|\nabla_{W_0}\|},\ 0.5,\ 2.0\right)
\]
Large gradient norm (saturated landscape) $\Rightarrow$ $\lambda < 1$ (contract).\\
Small gradient norm (flat landscape) $\Rightarrow$ $\lambda > 1$ (expand).

\vspace{0.3cm}
\textbf{Step 3 — Weight Update.}
The weight perturbation is the outer product of the scaled fixed point
with the attractor error, masked to the adjacency structure \cite{kosko1986,hebb1949}:
\[
\Delta W = \eta \cdot \lambda \cdot (\lambda H^*)^\top \otimes (H_{\text{target}} - H^*) \cdot M
\]
\[
W \leftarrow (W + \Delta W) \cdot M
\]
The update is accepted only if the reward $z = 2e^{-\|H_{\text{target}} - H_T\|} - 1$
strictly improves it; preserving the monotone guarantee of Jacobian gradient descent.

\vspace{0.3cm}
\textbf{Where:} \\
$\eta$ = the learning rate \\
$H$ = the state vector of the system \\
$H^*$ = the Newton fixed point of the attractor \\
$W$ = the FCM weight matrix \\
$\lambda$ = the adaptive spectral scale derived from gradient norm ratio \\
$E = H_{\text{target}} - H^*$ = the fixed-point error signal \\
$M$ = the structural adjacency mask \\
$\varepsilon$ = Newton convergence tolerance \\

\subsection*{Denoisification-Convergence Proof}

To establish denoising, assume the input is corrupted by additive noise \cite{bengio2013}:
\[
H_0 = H_{\text{clean}} + \varepsilon
\]

One aims to show:
\[
\|H_t - H_{\text{clean}}\| < \|H_0 - H_{\text{clean}}\|
\]

From the system dynamics:
\[
H_{t+1} = \phi(H_t)
\]

Consider the difference:
\[
\|\phi(H_t) - \phi(H_{\text{clean}})\|
\]

Using the Lipschitz property of the sigmoid \cite{lyapunov1992}:
\[
|\phi'(x)| \leq \frac{\lambda}{4}
\]

we obtain:
\[
\|\phi(H_1) - \phi(H_2)\| 
\leq \frac{\lambda}{4} \|(H_1 - H_2)(W + I)\|
\]

Bounding the equation one gets:
\[
\|\phi(H_1) - \phi(H_2)\| 
\leq \frac{\lambda}{4} \|H_1 - H_2\| \cdot \|W + I\|
\]

If:
\[
\frac{\lambda}{4} \|W + I\| < 1
\]

then there exists $\alpha \in (0,1)$ such that:
\[
\|\phi(H_1) - \phi(H_2)\| \leq \alpha \|H_1 - H_2\|
\]

Applying this to $H_t$ and $H_{\text{clean}}$:
\[
\|\phi(H_t) - \phi(H_{\text{clean}})\| \leq \alpha \|H_t - H_{\text{clean}}\|
\]

Assuming $H_{\text{clean}}$ is a fixed point:
\[
\phi(H_{\text{clean}}) = H_{\text{clean}}
\]

we obtain:
\[
\|H_{t+1} - H_{\text{clean}}\| \leq \alpha \|H_t - H_{\text{clean}}\|
\]

Iterating for $t \geq 1$:
\[
\|H_t - H_{\text{clean}}\| \leq \alpha^t \|H_0 - H_{\text{clean}}\|
\]

Since $0 < \alpha < 1$, it follows that:
\[
\|H_t - H_{\text{clean}}\| < \|H_0 - H_{\text{clean}}\|
\]
Thus, the error shrinks monotonically, hence the system exhibits denoising behavior.
\begin{figure}[htbp]
    \centering
    \includegraphics[width=0.6\textwidth]{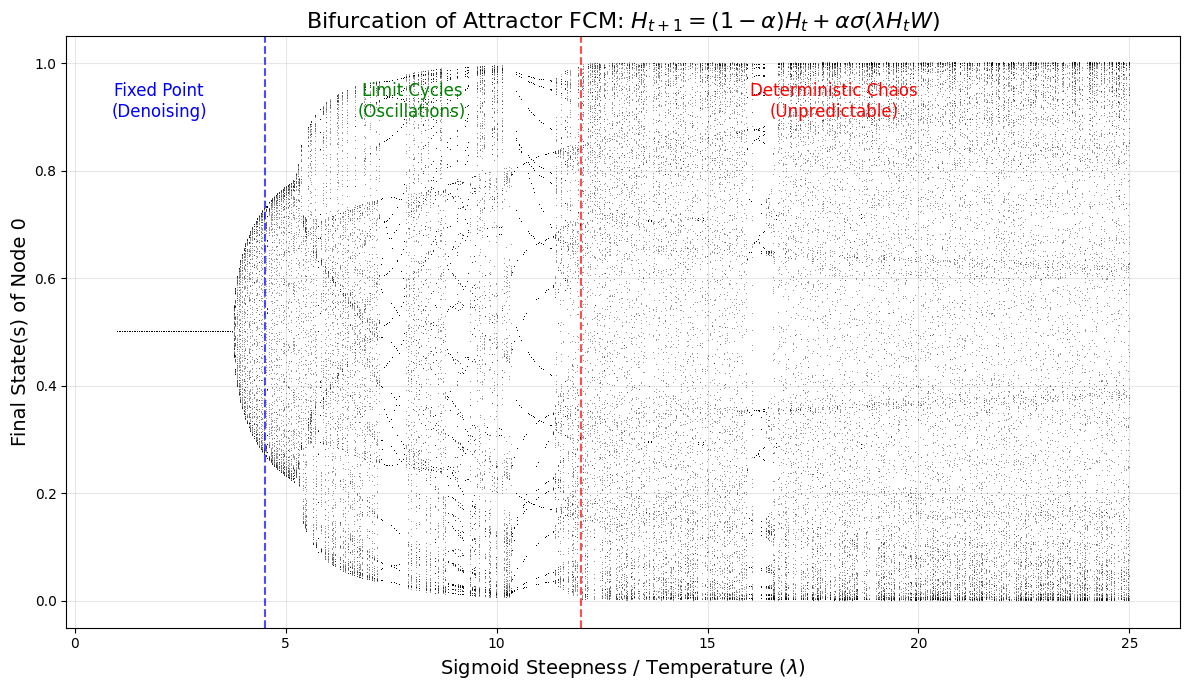}
    \caption{\textbf{Sigmoid Steepness Parameter $\lambda$:} Illustration of how the steepness affects the the denoisification behavior of the system.}
    \label{fig:steepness}
\end{figure}

\section{Discussion and Results}

The model is tested both quantitatively and qualitatively; there are three qualitative scenarios meant to push the model to it s limits. The point of the qualitative scenarios is to determine whether the model simulation results make sense. The attractor FCM is meant as a physics informed fuzzy cognitive map that respects the initial expert opinion. That can only be assessed by simulating relevant situations where the narrative before has to hold; in this case it the evaluation is purely empirical and focuses as to whether the outputs make sense. The scenarios are then illustrated in a map and the dynamics of the system are also plotted as they are a good reference point, for the attractor. 

\subsection*{Oligarchy Bailout, socioeconomic scenario}
In the first scenario, the model simulates an oligarchic society. Nodes 0-5 are the oligarchs while nodes 6-49 are the population; the model is then asked to redistribute more wealth from the population to the oligarchs. It is noted here that the population has to survive too, and that the wealth is limited. In other words, the problem is that the population does not have any more wealth. In a societal environment, other factors would have to be accounted for too. To address the issue, the model determined that the population would give all of their wealth to their oligarchs; as a corollary, the state eliminates weapons because the physics of that society would be violated. The population is poor, and hence the demand from the oligarchy cannot be achieved, but that doesn't mean it won't be attempted. Finally, according to the simulation, the population is simply done, which is a dark outcome, and due to the lack of resources, there cannot be a union at all.

\begin{figure}[htbp]
    \centering
    \includegraphics[width=\textwidth]{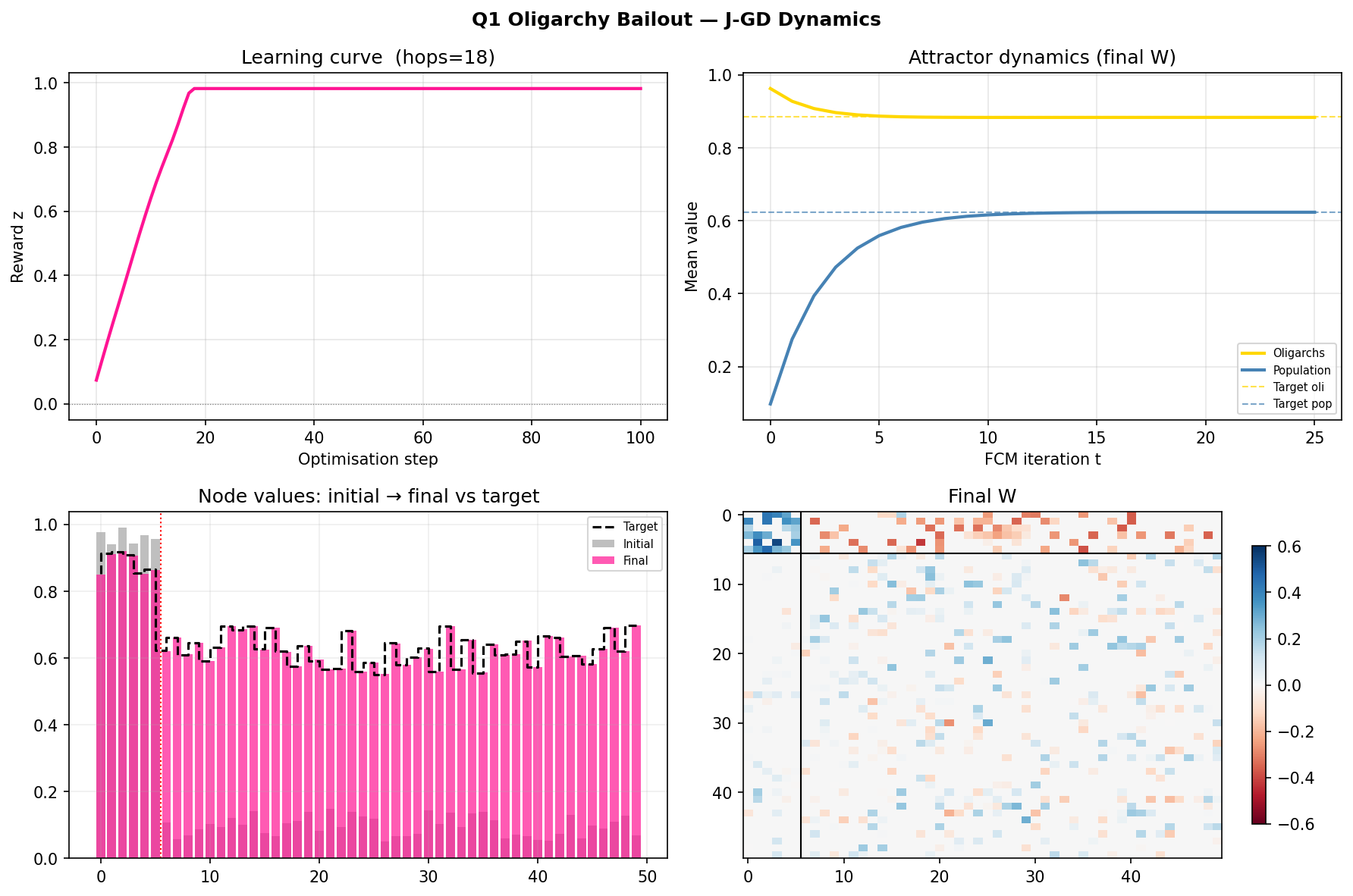}
    \caption{\textbf{Socio-Economic Stress Test:} The socioeconomic crisis.}
    \label{fig:oligarchy_sim}
\end{figure}
\subsection*{Ecological Incidence- ecological scenario}
In this scenario, the model simulates a trophic cascade; an ecological event occurs when the ecosystem is injected by a swarm of herbivores that forces the ecosystem into survival mode. For the ecological system the 0-3 nodes are for apex predators, 4-16 nodes are for herbivores, and 17-40 nodes are for producers. To resolve the issue of survival, a few things are predicted for the given ecosystem. The model determines that the predators will have to adapt and, therefore, hunt less to avoid overeating; the flora will also adapt by using defense mechanisms such as armor-toxicity. Producing more nutrition flowers is more like a death sentence for the flora, so flora's nutritional value will drop. Lastly, the herbivores will intake less calories from the flora until they will starve and the equilibrium target is achieved.
\begin{figure}[htbp]
    \centering
    \includegraphics[width=\textwidth]{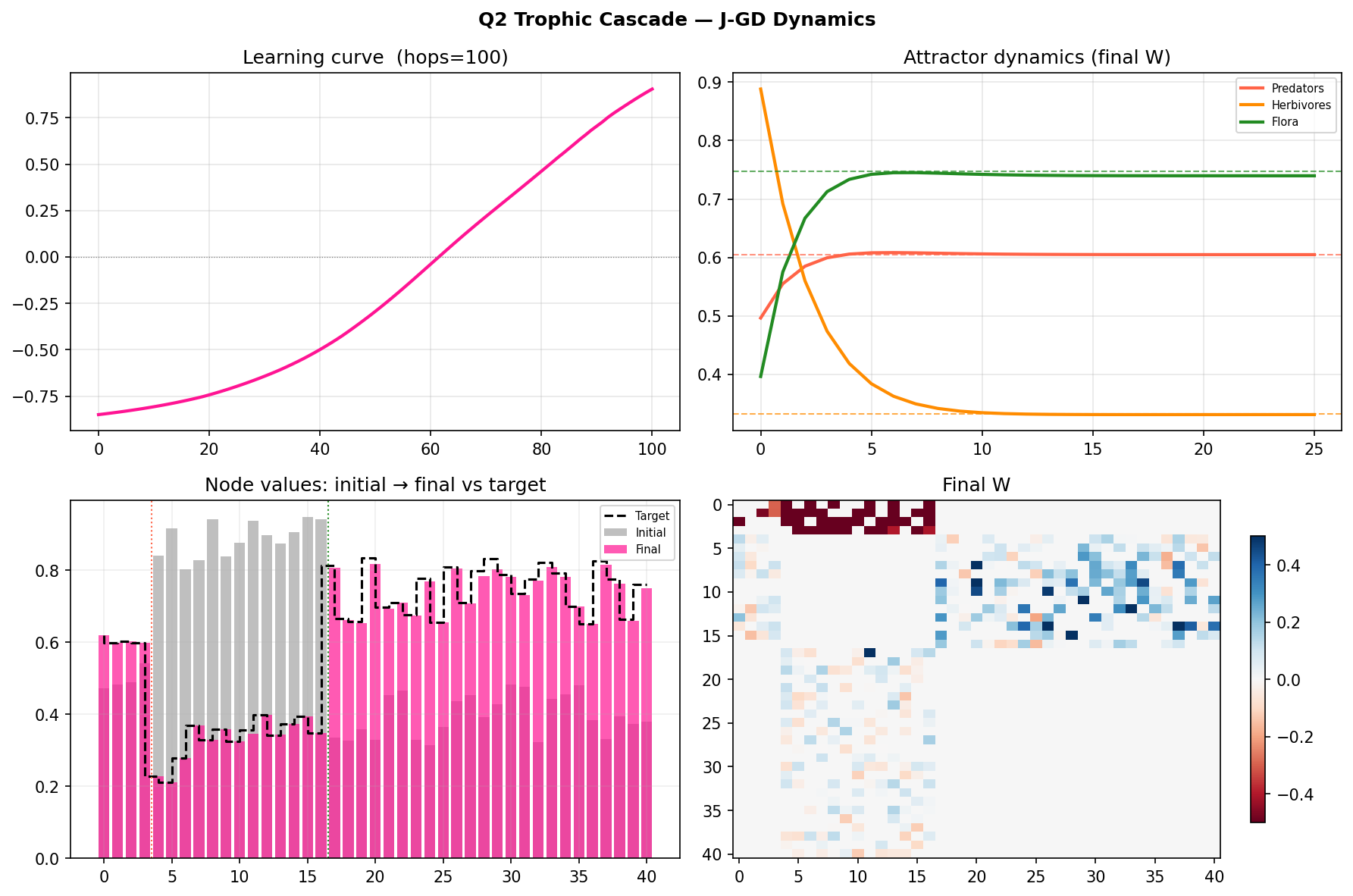}
    \caption{\textbf{Trophic Cascade Simulation:} survival strategies.}
    \label{fig:trophic_cascade}
\end{figure}
\subsection*{The Dictators Dilemma-political scenario}
In this scenario, the model simulates an authoritarian regime; a dictator wants to crush rioting from the dissidents. To address this setup, nodes 0-9 represent the regime facade, 10-25 nodes represent the dissidents in the regime, and 25-50 nodes represent the public. There is basically a riot, and the dictator wants to retain his grip over the population. The regime wants control of the situation all while a horde of dissidents is protesting for the regime. And that is probably the darkest scenario of all three, as according to the model, for the dictator to achieve a grip, he has to oppress the population further. To oppress the population further, he needs to eliminate the free speech, and to oppress the free speech, he must invent an authoritarian police force, effectively and literally eradicating the protest while doing so.
\begin{figure}[htbp]
    \centering
    \includegraphics[width=\textwidth]{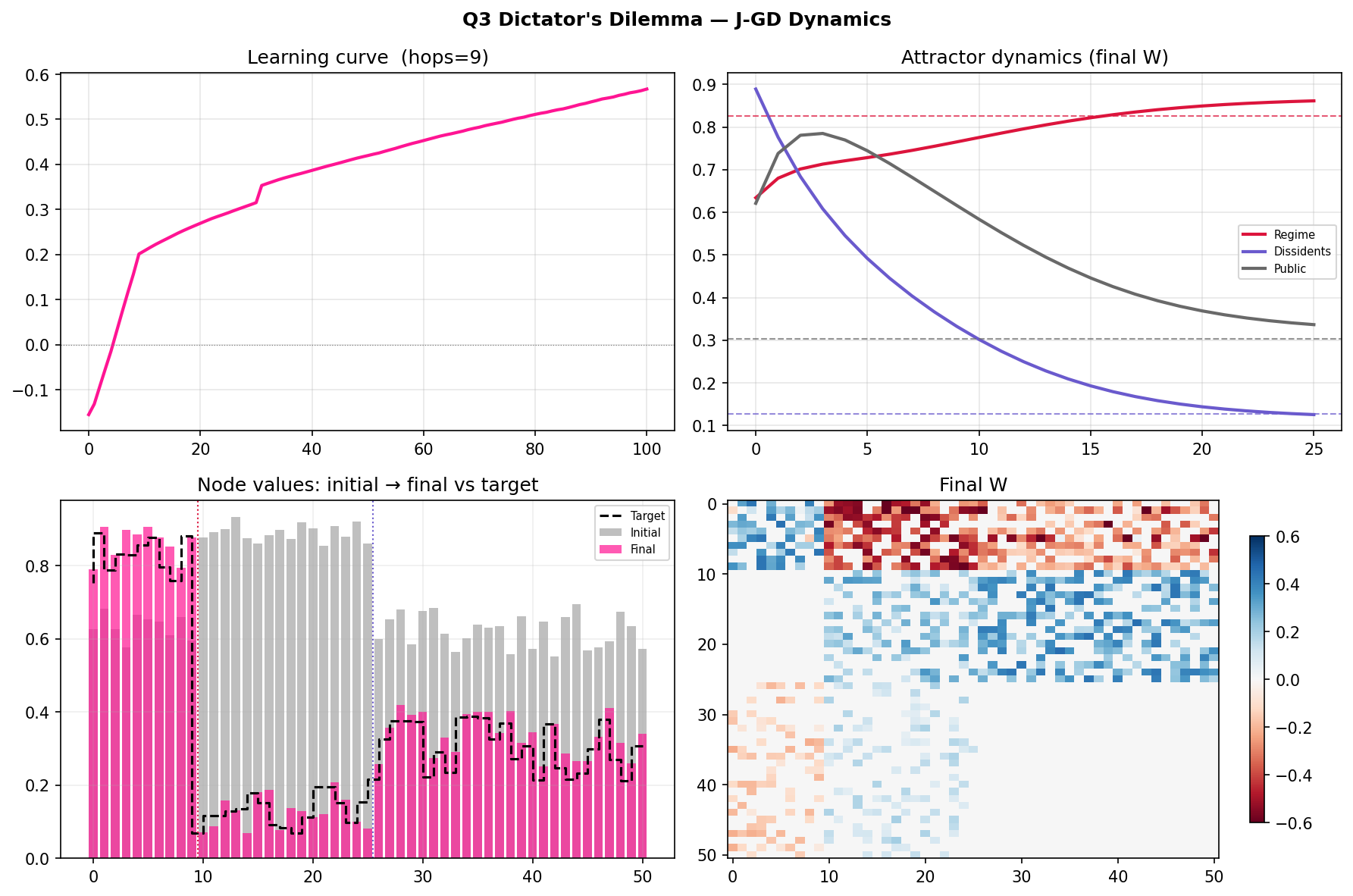}
    \caption{\textbf{Political Stress Test (The Dictator's Dilemma):} Comparison of state survival strategies. }
    \label{fig:dictator_dilemma}
\end{figure}
\pagebreak

The JGD point is to achieve lower energy state by finding the attractor and dynamically shifting the landscape to adjust learning rate correctly. To achieve that, an adaptive term that is directly linked to the weights that are calculated. The method then is evaluated for twenty fold through some qualitative scenarios and compared to other systems. For this a comparison between algorithms the following systems are used: simple, hebbian, agentic, Hybrid, and a custom FCM against a standard FCM. Testing the system thoroughly requires both quantitative and qualitative tests. There are seven tests in total: a stress test, a convergence test, a denoising test, a trap test, and the three scenarios.

\begin{table}[htbp]
\centering
\caption{Performance comparison across different scenarios. Lower values indicate better performance.}
\begin{tabular}{lccccccc}
\toprule
Scenario & Simple & Hebbian & ACO & GD-MCTS & GD & J-GD & Winner \\
\midrule
S1 Stress (n=20)      & 1.0008 & 1.3432 & 1.1886 & 1.2426 & 0.0586 & \underline{0.0211} & J-GD \\
S2 Convergence (n=20) & 1.0676 & 1.3033 & 1.2968 & 1.3289 & 0.0398 & \underline{0.0125} & J-GD \\
S3 Denoising (n=20)   & 1.4969 & 0.2782 & 0.6746 & 0.6822 & 0.0144 & \underline{0.0055} & J-GD \\
S4 Trap (n=20)        & 1.7016 & 0.7068 & 2.0040 & 1.9154 & 0.0490 & \underline{0.0057} & J-GD \\
Q1 Oligarchy (n=50)   & 1.1236 & 1.7115 & 0.4667 & 0.3951 & 0.0464 & \underline{0.0089} & J-GD \\
Q2 Trophic (n=41)     & 2.4562 & 2.1507 & 2.4696 & 2.4931 & \underline{0.0767} & 1.0397 & GD \\
Q3 Dictator (n=51)    & 0.7165 & 1.1721 & 0.7334 & 0.6322 & \underline{0.0726} & 0.3428 & GD \\
\bottomrule
\end{tabular}
\end{table}

From the results it is clear that the Jacobian gradient descent performs way better than any other FCM for simulating scenarios; it also has the interesting property of respecting the systems physics which make it a good choice. Some minor losses occur, however, by the gradient descent algorithm. That is probably due to the fact that the GD does not use a physics constrained mask and therefore it can concentrate solely on minimizing the target. Notice that the only in qualitative scenarios did the GD surpass the Attractor FCM, suggesting that the interpretation of such results might not be sound.
\subsection*{Ablation Study}
To verify that the system works and is cost efficient an ablation study is performed. First the mask is removed, then the residual memory, after that the JGD, after that the anchor, and lastly the system is striped to a basic fuzzy cognitive map configuration. It is noted that the pair combinations are also tested. Finally, the JGD is run purely without any additional modifications.
Following are the results of the ablation study:
\begin{table*}[ht]
\centering
\scriptsize
\begin{tabular}{lcccccccc}
\hline
Scenario & \shortstack{Full\\J-GD} & \shortstack{No\\Mask} & \shortstack{No\\Res.} & \shortstack{GD\\only} & \shortstack{No Mask\\+Res.} & \shortstack{No J-GD\\+Res.} & \shortstack{No J-GD\\+Mask} \\
\hline

S1 Stress (n=20)      
& 0.029 & 0.029 & \underline{0.022} & 0.982 & \underline{0.022} & 0.981 & 0.982 \\

S2 Convergence (n=20) 
& 0.022 & 0.022 & \underline{0.020} & 1.092 & \underline{0.020} & 1.092 & 1.092 \\

S3 Denoising (n=20)   
& \underline{0.005} & 0.005 & 0.005 & 0.027 & 0.005 & 0.027 & 0.027 \\

S4 Trap (n=20)        
& \underline{0.009} & 0.009 & 0.009 & 1.545 & 0.009 & 1.544 & 1.545 \\

Q1 Oligarchy (n=50)   
& 0.012 & 0.012 & \underline{0.012} & 0.259 & \underline{0.012} & 0.259 & 0.259 \\

Q2 Trophic (n=41)     
& 0.026 & 0.026 & \underline{0.023} & 2.365 & \underline{0.023} & 2.364 & 2.365 \\

Q3 Dictator (n=51)    
& 0.370 & 0.370 & \underline{0.112} & 1.499 & \underline{0.112} & 1.448 & 1.499 \\

\hline
\end{tabular}
\caption{Performance comparison (Part 1). Best values per row are underlined (ties included).}
\end{table*}

\begin{table*}[ht]
\centering
\scriptsize
\begin{tabular}{lcccccc}
\hline
Scenario & \shortstack{No\\Anchor} & \shortstack{No Anchor\\+Mask} & \shortstack{No Anchor\\+Res.} & \shortstack{No Anchor\\+J-GD} & \shortstack{No Anchor\\+Mask+Res.} & \shortstack{Simple\\FCM} \\
\hline

S1 Stress (n=20)      
& 0.168 & 0.168 & 0.165 & 0.709 & 0.165 & 1.001 \\

S2 Convergence (n=20) 
& 0.247 & 0.247 & 0.250 & 0.856 & 0.250 & 1.068 \\

S3 Denoising (n=20)   
& 0.018 & 0.018 & 0.018 & 0.436 & 0.018 & 1.497 \\

S4 Trap (n=20)        
& 0.042 & 0.042 & 0.042 & 1.192 & 0.042 & 1.702 \\

Q1 Oligarchy (n=50)   
& 0.175 & 0.175 & 0.175 & 0.181 & 0.175 & 1.124 \\

Q2 Trophic (n=41)     
& 0.200 & 0.200 & 0.200 & 2.005 & 0.200 & 2.456 \\

Q3 Dictator (n=51)    
& 0.358 & 0.358 & 0.350 & 0.362 & 0.350 & 0.717 \\

\hline
\end{tabular}
\caption{Performance comparison (Part 2).}
\end{table*}
Whatever the case, the program might overshoot a bit, and it is slower to converge due to the additional anchor $H_t$ term. Furthermore, it is computationally more expensive, as it constantly needs to perturb the landscape for the adaptive term to correctly navigate the descent. However, the attractor FCM provides both qualitative and quantitative results that are great. That stated, the potential of the system should not be underestimated. An interesting thing has to do with the holistic approach; some versions seem to perform better under other scenarios, while others don't. The point of this system was to find an interpretable, and overall high performing system, regardless of the computational complexity overhead.

Not only the full system performs predictably well under all circumstances, and predictability matters. Nevertheless, what truly makes it good is the JGD learning program. Removing the JGD resulted in generally declined performance among all stress tests and qualitative scenarios. What is more impressive is that the mask did not necessarily affect the results positively. But that probably occurs due to the Jacobian gradient descent nature and the current weights, meaning the mask was already placed by the Jacobian gradient descent program. Lastly, what is really striking is that for the working system major adjustments result in minor improvements. The system is already saturated, and despite adding more quirks the improvements are minor; the best system was a simple FCM with an anchor for which the JGD algorithm applied better, eventually.

\section{Conclusion}
To conclude an ablation study was done and the system works best without residuals and a mask; along with some stress test that show that the system works as intended, both the quantitative and the qualitative. Using the Newton's attractor dynamics with gradient descent seems a great way to optimize fuzzy cognitive maps learning. The adjacency mask constrain also make it an interpretable system that imputes the physical meaning, which aligns closely with the experts. In the three qualitative scenarios that were tested such a system resulted in a hard truth about the matter.

Furthermore, there can be other uses as well such as denoisification, as proven. In other words, the attractor FCM can accept a noisy input and correctly guide the target closer to the expert input. Despite the noise, the user input will reach the correct target and that is absolute. Finally, the model will always converge a fixed point. Reaching a fixed point, means it will produce a meaningful and valid result.

\pagebreak
\section*{Acknowledgments}
It is acknowledged that this paper is part of a PhD dissertation, fuzzy optimization of information transmission in service design process currently done in (Athens University of Economics and Business) AUEB. It is also written in correspondence with Dr. Dimitris Kardaras which was the supervisor of the specific subject in AUEB and also interested in service optimization.

\end{document}